\documentclass[letterpaper]{article} 
\usepackage{aaai25}  
\usepackage{times}  
\usepackage{helvet}  
\usepackage{courier}  
\usepackage[hyphens]{url}  
\usepackage{graphicx} 
\usepackage{natbib}  
\usepackage{caption} 
\usepackage{booktabs}  

\usepackage{subfig}
\usepackage{enumitem}
\usepackage{algorithm}
\usepackage{algorithmic}
\usepackage{makecell}
\usepackage{amsthm}
\usepackage{amsmath} 
\usepackage{amsfonts} 
\usepackage{amssymb} 
\usepackage{graphicx} 
\usepackage{bm} 

\usepackage{xcolor}

\usepackage{newfloat}
\usepackage{listings}
\DeclareCaptionStyle{ruled}{labelfont=normalfont,labelsep=colon,strut=off} 
\floatstyle{ruled}
\newfloat{listing}{tb}{lst}{}
\floatname{listing}{Listing}
\title{POPoS: Improving Efficient and Robust Facial Landmark Detection with Parallel Optimal Position Search}

\author{
    Chong-Yang Xiang\textsuperscript{1}\thanks{Partially completed during a visit to Alibaba and CMU.}, 
    Jun-Yan He\textsuperscript{2}, 
    Zhi-Qi Cheng\textsuperscript{3}\thanks{Corresponding author: Zhi-Qi Cheng (zhiqics@uw.edu). Partly completed as a Project Scientist at CMU.}, 
    Xiao Wu\textsuperscript{1}, 
    Xian-Sheng Hua\textsuperscript{4}
}

\affiliations{
    \textsuperscript{1}School of Computing and Artificial Intelligence, Southwest Jiaotong University, Chengdu, China \\
    \textsuperscript{2}Institute for Intelligent Computing, Alibaba Group, Shenzhen, China \\
    \textsuperscript{3}School of Engineering and Technology, University of Washington, Tacoma, WA, USA \\
    \textsuperscript{4}College of Computer Science and Technology, Zhejiang University, Hangzhou, China \\
}

\usepackage{bibentry}

\begin{document}

\maketitle

\begin{abstract}
Achieving a balance between accuracy and efficiency is a critical challenge in facial landmark detection (FLD). This paper introduces Parallel Optimal Position Search (POPoS), a high-precision encoding-decoding framework designed to address the limitations of traditional FLD methods. POPoS employs three key contributions: (1) Pseudo-range multilateration is utilized to correct heatmap errors, improving landmark localization accuracy. By integrating multiple anchor points, it reduces the impact of individual heatmap inaccuracies, leading to robust overall positioning. (2) To enhance the pseudo-range accuracy of selected anchor points, a new loss function, named multilateration anchor loss, is proposed. This loss function enhances the accuracy of the distance map, mitigates the risk of local optima, and ensures optimal solutions. (3) A single-step parallel computation algorithm is introduced, boosting computational efficiency and reducing processing time. Extensive evaluations across five benchmark datasets demonstrate that POPoS consistently outperforms existing methods, particularly excelling in low-resolution heatmaps scenarios with minimal computational overhead. These advantages make POPoS as a highly efficient and accurate tool for FLD, with broad applicability in real-world scenarios. The code is available at \url{https://github.com/teslatasy/POPoS}.
\end{abstract}

\section{Introduction}
\label{sec:intro}

Facial Landmark Detection (FLD) is a fundamental task in computer vision, essential for identifying and localizing key facial features such as the eyes, nose, and mouth in digital images~\cite{wu2018facial}. This crucial process underpins a wide array of advanced applications, including facial recognition systems~\cite{khabarlak2021fast} and three-dimensional face modeling~\cite{zeng2023d}. Beyond mere detection, FLD serves as a cornerstone for interpreting human facial expressions~\cite{cheng2024mips,cheng2024sztu}, analyzing emotional states~\cite{cheng2024emotion}, and facilitating interactions within digital environments~\cite{cheng2016video,cheng2017video,cheng2017video2shop,cheng2017selection}. Consequently, FLD drives innovation across diverse fields such as human-computer interaction~\cite{li2024human,he2023wordart,he2024metadesigner,he2023wordart} and augmented reality systems~\cite{xu2024facechain,tu2024motioneditor,tu2024motionfollower,xu2024combo}.

The advent of deep learning has advanced FLD research, primarily through two dominant approaches: heatmap-based methods and coordinate regression techniques. Heatmap regression methods~\cite{Chandran_attention_CVPR2020,Zhang_darkpose_CVPR20,Jin_PIPNet_IJCV21} have garnered considerable attention due to their ability to generate probabilistic heatmaps for individual landmarks. These methods effectively preserve spatial and contextual information while mitigating the risk of overfitting during training. Additionally, they adeptly capture the inherent uncertainty in landmark positions, enabling robust performance across a wide range of facial expressions and orientations.

\begin{figure}[t]
  \centering
   \includegraphics[width=0.95\linewidth]{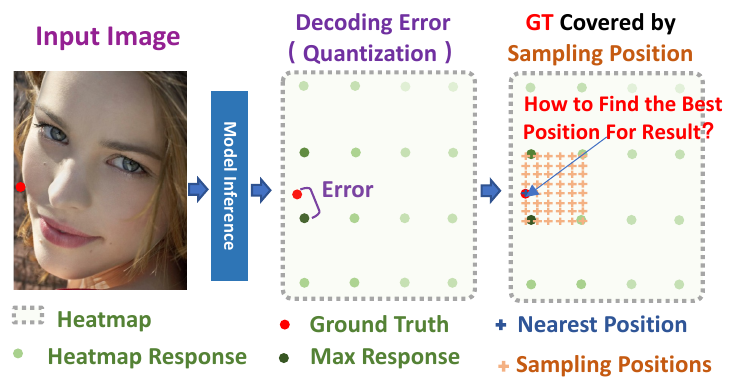}
   \vspace{-1em}
    \caption{Quantization error in the coordinate decoding process. The discrepancy between the ground truth (GT) coordinates and the highest probability point in the heatmap illustrates the challenge of achieving precise localization and lossless decoding.~[Best viewed in color with zoom]} 
   \vspace{-2em}
   \label{fig:intro}
\end{figure}

Despite their strengths, heatmap-based approaches encounter significant challenges. A primary limitation is quantization error, which becomes pronounced when downscaling heatmaps from their original resolution~\cite{bao2023keyposs}. This issue stems from the discretization of continuous spatial information into pixelated representations, leading to potential precision loss during landmark localization. Although maintaining high-resolution heatmaps can mitigate this problem, it imposes substantial computational demands, rendering it impractical for real-time applications or deployment on resource-constrained devices.

Furthermore, the core challenge in FLD lies in balancing localization accuracy with computational efficiency, particularly in scenarios that require real-time performance or operate under limited processing power. This trade-off is increasingly critical as applications demand both high precision and speed . While pioneering works such as Hourglass~\cite{Newell_Hourglass_ECCV16}, CrossNet~\cite{zhang2022crossnet}, and KeyposS~\cite{bao2023keyposs} have made notable strides in reducing accuracy degradation in low-resolution heatmaps, achieving an optimal balance between accuracy and efficiency remains an open challenge. Current state-of-the-art methods often struggle to maintain high accuracy under low-resolution conditions or stringent computational constraints, indicating a need for further exploration.

This ongoing pursuit for balance in FLD is compounded by the inherent limitations of current encoding and decoding techniques, which further restrict localization performance. Research efforts are actively exploring ways to enhance heatmap-based approaches by leveraging their strengths while mitigating their weaknesses, often through hybrid architectures or advanced post-processing strategies.
Fig.~\ref{fig:intro} illustrates a key challenge in FLD: quantization error in coordinate decoding. The figure highlights the misalignment between the ground truth (GT) landmark coordinates and the heatmap's highest-probability point. This discrepancy arises from the discretization process, where the continuous spatial domain is mapped onto a discrete pixel grid, compromising localization accuracy. The visualization underscores the importance of advanced decoding strategies to bridge the gap between discrete heatmap representations and the continuous spatial domain of facial features.
These limitations emphasize the pressing need for innovative methods to improve accuracy while addressing the challenges of discretization and quantization errors.

To address these interconnected challenges, we present the \textit{Parallel Optimal Position Search (POPoS)} framework, a novel solution designed to resolve the accuracy-efficiency trade-offs inherent in heatmap-based methods. POPoS introduces a set of complementary strategies to tackle the core limitations of traditional FLD approaches. At its foundation, POPoS employs a pseudo-range multilateration technique that leverages the top $K$ points of the heatmap as anchors, effectively reducing the influence of multiple peaks and enhancing landmark localization accuracy. This is further complemented by an anchor-based optimization strategy that integrates global and local refinements, compensating for quantization errors and achieving sub-pixel precision.
In addition, POPoS incorporates a single-step parallel computation algorithm, significantly reducing computation time while preserving high accuracy—an approach well-suited to modern GPU architectures. These improvements enable POPoS to excel in low-resolution heatmap scenarios, striking an effective balance between precision and computational efficiency. Consequently, POPoS emerges as a versatile solution applicable to a wide range of real-world settings, from mobile platforms to real-time systems.
The key contributions of POPoS are summarized as follows:

\begin{itemize}
    \item A pseudo-range multilateration approach that improves localization in scenarios with multiple local maxima or ambiguous heatmap regions.
    \item An anchor-based optimization strategy that synergistically balances global and local optimization, enhancing sub-pixel accuracy.
    \item A GPU-optimized, single-step parallel computation algorithm that significantly improves efficiency over traditional iterative methods.
    \item Extensive evaluation across multiple datasets demonstrating superior performance in low-resolution heatmap scenarios, with minimal computational overhead.
\end{itemize}

\section{Related Work}
\label{sec:related_work}
Facial landmark detection research has two key directions: heatmap regression and high-efficiency heatmap decoding.

\subsection{Heatmap Regression-Based Methods}
Heatmap regression methods~\cite{Chandran_attention_CVPR2020,Sun_HRNet_CVPR19,Tang_Quan_ECCV18,Zou_Robust_ICCV19,Wang_AWingLoss_ICCV19,Jin_PIPNet_IJCV21,cheng2019improving,cheng2019learning,cheng2022rethinking,huang2020stacked,zhang2022crossnet} have progressed by predicting landmarks using high-resolution feature maps. The Stacked Hourglass Network~\cite{Newell_Hourglass_ECCV16} and UNet~\cite{Ronneberger_UNet_MICCAI15} established the foundation by preserving spatial relationships. Subsequent innovations addressed specific challenges: UDP~\cite{Huang2020UDPpose} tackled discretization errors, G-RMI~\cite{Papandreou2017G-RMI} mitigated quantization errors by combining dense heatmaps with keypoint offsets, and the DSNT layer~\cite{Nibali2018Numerical} introduced coordinate supervision regularization. PIPNet~\cite{Jin_PIPNet_IJCV21} further streamlined the process with simultaneous predictions on low-resolution heatmaps. This progression has consistently enhanced accuracy by leveraging spatial pixel relationships~\cite{Zou_Robust_ICCV19,Wang_AWingLoss_ICCV19}.

\subsection{High Efficiency Heatmap Decoding Methods}
While heatmap regression improved prediction accuracy, extracting coordinates from estimated heatmaps remained a critical area for optimization. High-efficiency decoding methods were developed to address this challenge. Darkpose~\cite{Zhang_darkpose_CVPR20} refined heatmap generation and distribution analysis, while FHR~\cite{Stable_Face_Alignment_AAAI19} enhanced precision through improved fractional part estimation. Subpixel heatmap regression~\cite{Bulat2021Subpixel} increased accuracy by predicting offsets using soft continuous maxima. KeyPosS~\cite{bao2023keyposs} introduced a novel approach by integrating GPS-inspired multi-point positioning technology for true-range estimation.

Building on these advancements, our Parallel Optimal Position Search (POPoS) method combines pseudo-range multilateration and Potential Position Parallel Sampling. POPoS addresses limitations in predictive accuracy and computational efficiency, particularly in scenarios with low-resolution heatmaps. This innovation not only advances facial landmark detection but also extends its applicability to resource-constrained environments. It marks a step towards more accessible and efficient high-accuracy detection across diverse applications and hardware configurations.

\section{Parallel Optimal Position Search (POPoS)}
\label{sec:proposed_method}
\begin{figure*}[t]
  \centering
   \includegraphics[width=0.95\linewidth]{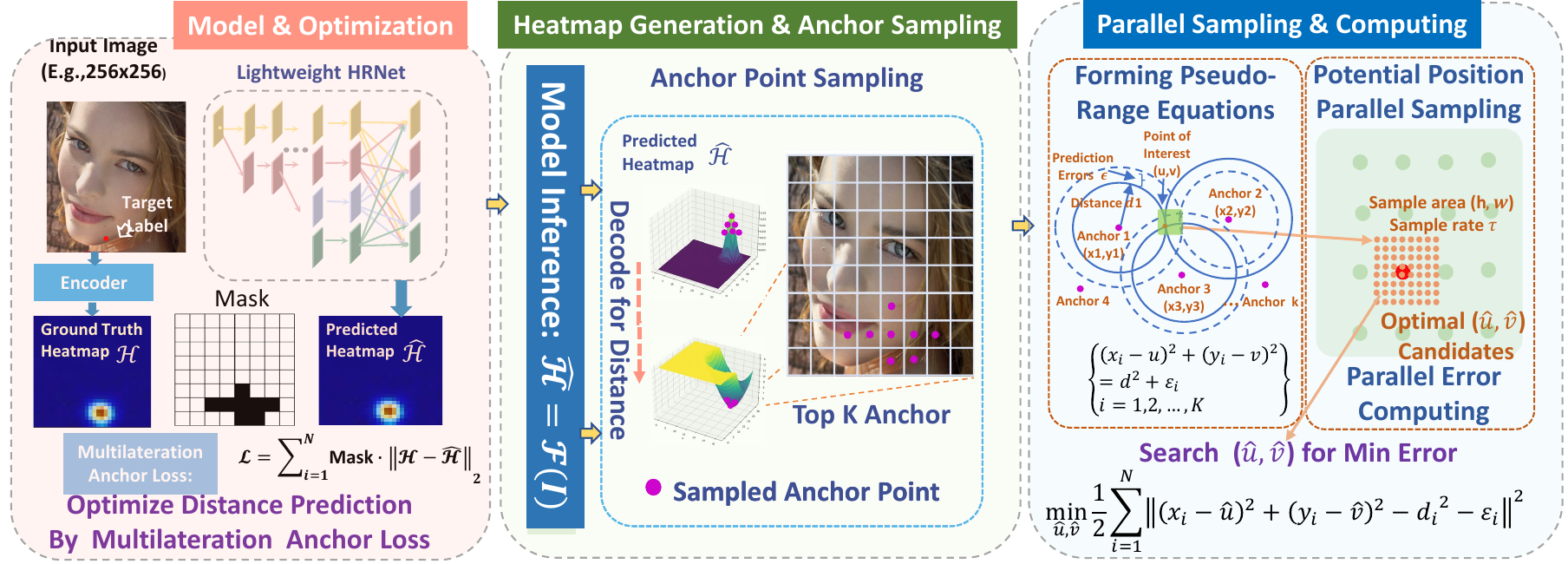}
   \vspace{-0.5em}
   \caption{Parallel Optimal Position Search (POPoS) framework. The Multilateration Anchor Loss is designed during the training phase to optimize distance prediction. During model inference, POPoS samples the top-K response positions as anchor points for distance map decoding.~[Best viewed in color with zoom]}
   \label{fig:framwork}
   \vspace{-1.5em}
\end{figure*}

Despite advancements like HRNet~\cite{Sun_HRNet_CVPR19}, heatmap-based facial landmark detection faces persistent challenges. Given an input image $I \in \mathbb{R}^{H \times W \times 3}$, the objective is to locate $N_k$ facial landmarks $\{\boldsymbol{\beta}_{k}\}_{k=1}^{N_k}$, where $\boldsymbol{\beta}_k = (u_k, v_k)$. Current approaches use CNNs to generate heatmaps $\mathcal{H} \in \mathbb{R}^{h \times w \times N_k}$, with $h = H / \lambda$, $w = W / \lambda$, and $\lambda$ as the downsampling factor. Key challenges include:

\begin{itemize}
    \item \textbf{Resolution-Efficiency~Trade-off}.~High-resolution heatmaps ($\lambda \to 1$) provide precise localization but increase computational load, hindering real-time performance~\cite{Newell_Hourglass_ECCV16}. Low-resolution heatmaps ($\lambda \gg 1$) are efficient but less accurate, impacting applications in resource-constrained scenarios.
    \item \textbf{Encoding-Decoding Dilemma}.~Biased encoding ($\boldsymbol{\beta}_{\text{biased}} = \text{Quan}(\boldsymbol{\beta}/\lambda)$) introduces quantization errors, especially at large $\lambda$~\cite{Zhang_darkpose_CVPR20}. Unbiased encoding ($\boldsymbol{\beta}_{\text{unbiased}} = \boldsymbol{\beta}/\lambda$) maintains accuracy but results in diffuse heatmaps and increased computational cost~\cite{Zhang_darkpose_CVPR20}.
    \item \textbf{Decoding Limitations}.~One-hot decoding ($\hat{\boldsymbol{\beta}}=\operatorname{argmax}(\mathcal{H})$) is efficient but prone to discretization errors.
    Distribution-aware methods~\cite{Zhang_darkpose_CVPR20} perform well at high resolutions but struggle with low-resolution or multi-modal heatmaps, particularly in complex scenarios~\cite{Chandran_attention_CVPR2020}.
    \item \textbf{Optimization Constraints}.~The commonly used Mean Squared Error (MSE) loss assumes Gaussian error distribution, often inappropriate for facial landmarks~\cite{Zhang_darkpose_CVPR20}. It fails to capture spatial relationships between landmarks, leading to anatomically implausible configurations.
\end{itemize}
These interconnected challenges hinder robust, accurate, and efficient facial landmark detection in a wide range of scenarios.~To address these challenges, we propose Parallel Optimal Position Search (POPoS), a framework that fundamentally reimagines heatmap-based facial landmark detection. POPoS comprises three stages, as shown in Fig.~\ref{fig:framwork}. First, the model optimization phase involves generating heatmaps and optimizing them, where the multilateration anchor loss is applied to minimize encoding information loss. Second, the heatmap generation and anchor sampling phase include model inference, distance transformation, and anchor point selection. Third, the sample and search phase involves constructing the pseudo-range positioning equation, parallel sampling, and equivalent single-step parallel computation. 
Details are in the following sections.

\subsection{Model \& Optimization}
Our adopted HRNet \cite{bao2023keyposs}, specially optimized for memory efficiency, is particularly suitable for mobile and edge computing applications. Its optimizations include early downsampling and a consistent application of 3×3 convolutional kernels, ensuring a balance between performance and computational efficiency.
The optimization objective of current Mean Squared Error (MSE) loss is to minimize the average squared difference between the predicted values and the actual values of the entire heatmap. While the predicted values often biased towards outliers at low resolutions after optimization, which will induce encoding errors. To enhance accuracy of the distance map for anchor points during the decoding step, the inconsistent optimization objectives between the decoding and encoding steps need to be solved. 
The Multilateration Anchor (MA) loss (Eq.\ref{eq:local anchor loss}) is proposed to address the optimization challenge by enhancing accuracy within the top K points, where top K points correspond to the K station anchors at the decoding step:
\begin{equation}\label{eq:local anchor loss}
\mathcal{L}_\text{MA} = \sum_{i=1}^{N_k} \|\mathcal{H}_{i} \cdot \text{Mask}_{i} - \hat{\mathcal{H}}_{i} \cdot \text{Mask}_{i} \|_{2},
\end{equation}
where $\mathcal{H}_{i}$ and $\hat{\mathcal{H}}_{i}$ are the ground truth and predicted heatmaps, and $\text{Mask}_{i} \in R^{h\times w}$ indicates a binary mask which assigns the top K response locations with value one, while the others remain zero, $i$ represents the $i$-th landmark.
Further, the model is jointly optimized using MSE loss and MA loss. The loss $L$ in the training stage is $\mathcal{L}=\mathcal{L}_\text{MSE}+{w} \times \mathcal{L}_\text{MA}$, where $\mathcal{L_\text{MSE}}$ is the MSE loss of the whole heatmap, $\mathcal{L}_\text{MA}$ is the MA loss, and ${w}$ is the weight parameter of local anchor loss.
\subsection{Heatmap Generation \& Anchor Sampling}
In this stage, there are three key steps:1)~heatmap generation, 2)~distance transformation, and 3)~anchor sampling.

First step is heatmap generation. An input image $I \in \mathbb{R}^{H \times W \times 3}$ is processed through a streamlined HRNet variant~\cite{bao2023keyposs} to extract features. The heatmaps $\mathcal{H} \in \mathbb{R}^{h \times w \times N_k}$ is then generated, where $h = H / \lambda$, $w = W / \lambda$, and $N_k$ represents the number of facial landmarks.
The unbiased coordination encoding is employed during the encoding stage. The coordinate heatmap is generated as a 2-dimensional Gaussian distribution/kernel centered at the labeled coordinate.  

The second step is distance transformation, which converts the heatmap into a distance map. The distance between the predicted position and each heatmap pixel, $\mathcal{D}(i,j)$, is calculated as $\mathcal{D}(i,j) = \sqrt{-2\sigma^{2} \ln(\mathcal{H}(i,j))}$ , where $\sigma$ is the standard deviation of the Gaussian distribution. {Because the heatmap retains the initial positional information entirely, thereby the distance heatmap maintains sub-pixel accuracy.}

The third step is anchor sampling.~To mitigate the prediction errors induced by heatmap generation, different from the triangulation algorithm~\cite{Stable_Face_Alignment_AAAI19}, the top K points of the distance map are selected as anchors. Because the selected top K points are consistent with those top K points in the model optimization stage, they promise to minimize information loss. Meanwhile, these selected top K points are utilized to form distance equations in the next stage.

\subsection{Parallel Sampling \& Computing}
This stage involve two key steps:~1)~forming the pseudo-range multilateration equation, and~2)~potential position parallel sampling and computing as follows:

\subsubsection{Forming Pseudo-range Multilateration Equations} The distance map decoding is treated as the pseudo-range multilateration (PRM) problem.  Here we utilize selected top K points as anchors, to form a set of distance equations. These distance equations represent the distances between multiple station anchors with the target point. Formally, the top K anchors are $\mathcal{A}=\{A_{(x_{0},y_{0})}, \ldots, A_{(x_{K},y_{K})}\}$, the real distances  between $\mathcal{A}$ and the real poistion point of interest {$\boldsymbol{\beta}({u},{v})$ } are $\mathcal{D}=\{d_{0}, \ldots, d_{K}\}$, and the prediction distance error for each anchor is denoted by $\varepsilon_i$. The system of non-linear PRM equations is given by Eq.~\ref{eq:non-linear}:

{\small
\begin{equation}\left\{(x_i-u)^2+(y_i-v)^2=d_i^2+\varepsilon_i\mid i=0,1,\ldots,K\right\}.
    \label{eq:non-linear} 
\end{equation} 
}

\subsubsection{Potential Position Parallel Sampling and Computing} 
To solve the pseudo-range multilateration equation and find the point of interest $\boldsymbol\hat{\beta}=(\hat{u},\hat{v})$, the objective function is constructed as Eq.~\ref{eq:objective}:
    \begin{equation}
    \min_{\hat{u},\hat{v}}\frac{1}{2}\sum_{i=1}^{K}\left\|\left(x_i-\hat{u}\right)^2+\left(y_i-\hat{v}\right)^2-d_i^2-\varepsilon_{i}\right\|^2,
    \label{eq:objective}
    \end{equation}
where $\varepsilon_{i}$ represents the error between the decoded distance and the actual distance. A new method for solving the objective function is proposed, called Potential Position Parallel Sampling and Computing (PPPSC). Through three steps (potential region selection, candidate generation, and Solution of Parallel Computing), it can directly find the point that minimizes the objective function. The detailed steps are outlined and further explained below.

\noindent \textit{(1)~Potential Region Selection.}~First, considering the subpixel scenario, there are an infinite number of points existing in the heatmap. Second, as the distance between the sample point with the maximum value point $m$ in the heatmap increases, the probability of samping point hitting the point of interest decreases. Third, an excess of sampling can lead to redundant computations. Hence, the search area is set as a square region of $h\times w$ centered at point $m$.

\noindent \textit{(2)~Candidates Generation.}~Within the selected region, sampling is conducted using a mesh with a frequency of $\tau$ samples per pixel and obtained a total of $N$ points. Formally, a mesh grid point set $G=\{g_{1}, \dots, g_{N}\}$ is generated first, $g_{i}=(\frac{\Delta\boldsymbol\hat{u}_{i}}{\lambda }, \frac{\Delta\boldsymbol\hat{v}_{i}}{\lambda })$,
    where $\Delta\boldsymbol\hat{u}_{i} \in [0, \lambda \times h \times \tau], \Delta\boldsymbol\hat{v}_{i} \in [0, \lambda \times w \times \tau]$. And $\tau$ is a factor to control the granularity of subpixel sampling. $\lambda$ is the downsampling factor. Then, the mesh grid point set $G$ is translated to cover the maximum value position $(\hat{u}_{m}, \hat{v}_{m})~$in the grid center, and the candidates $c_{i} \in C$ is computed as Eq.~\ref{eq:ci}:
    \begin{equation}
         c_{i}(\hat{u}_{i},\hat{v}_{i})=(\frac{\Delta\boldsymbol\hat{u}_{i}}{\lambda  } + \hat{u}_{m} -\frac{w}{2},  \frac{\Delta\boldsymbol\hat{v}_{i}}{\lambda  }+ \hat{v}_{m} -\frac{h}{2}).
         \label{eq:ci}
    \end{equation}

\noindent \textit{(3)~Solution of Parallel Computing.}~After the sampling, the computing step for one heatmap is formed as Eq. \ref{eq:objective_sample}. All points from the sampling set $C$ are then inputted into the objective function $\Phi(\mathcal{A}, C)$, which calculates the discrepancy between two distance matrices: one is the Euclidean distance matrix between the sampled points $C$ and the anchor set $\mathcal{A}$, and the other one is the predicted distance map $D$ of anchor set $\mathcal{A}$. This results in an error matrix \( E \). 
The optimal facial landmark position is determined by locating the minimum position index in the matrix \( E \).
    \begin{align}
        &\Phi(\mathcal{A}, C) = \arg\min_i \sum_k (||C-A||_2-D)_{ik} \notag \\
                   &= \min_{\hat{u},\hat{v}} \sum_{i=1}^{K} \left\| \left(x_i - \hat{u}\right)^2 + \left(y_i - \hat{v}\right)^2 - d_i^2 - \varepsilon_{i} \right\|^2,
        \label{eq:objective_sample}
    \end{align}
where $x_i$ and $y_i$ are the coordinates of the $i$-th anchor point. $\hat{u}$ and $\hat{v}$ are the coordinates of the sampled point, $d_i$ is the measured distance from anchor $i$ to the target point, $\varepsilon_i$ represents error between the decoded distance and the actual distance for the $i$-th anchor.
The optimization objectives of Eq.~\ref{eq:objective} and Eq.~\ref{eq:objective_sample} are consistent, thus PPPSC methods can achieve the optimal approximate solution.

At last, Iterative Gauss–Newton Optimization (IGNO) for PRM is introduced. To validate the effectiveness of our approach, we implemented a general method, IGNO, for solving nonlinear equations of the PRM. Specifically, we utilized the Gauss-Newton iteration method to solve the PRM objective function Eq.~\ref{eq:objective}. To find the point of interest $\boldsymbol\hat{\beta}=(\hat{u},\hat{v})$,
    {the iteration starts at some initial guess $\boldsymbol\hat{\beta}_0$, and update }$\boldsymbol\hat{\beta}_{n+1}=\boldsymbol\hat{\beta}_n+\Delta\boldsymbol\hat{\beta}$. At each iteration, the update $\Delta\boldsymbol\hat{\beta}$ can be calculated from Eq.~\ref{eq:delta}:
    \begin{equation}
    \left(\sum_{i=1}^{k}\boldsymbol{J}_i^{-1}\boldsymbol{J}_i^\mathrm{T}\right)\Delta\boldsymbol\hat{\beta}=\sum_{i=1}^{K}-\boldsymbol{J}_i^{-1}e_i,
    \label{eq:delta}
    \end{equation} 
where $k$ represents the number of anchors, $J$ is Jacobian matrix of the objective function, $e_i$ represents the objective function value at the i-th iteration. The iteration process halts either when $e_i$ gradually converges or when it reaches the upper limit of iterations. The IGNO methods require computations for each point individually, which has some challenges such as extensive loop operations and GPU acceleration. The proposed PPPSC approach only involves matrix operations, and is thus inherently suitable for parallel computation on GPUs. The computational complexity of the PPPSC approach can be expressed as $4 \times N_a \times N_c$, where $N_a$ represents the number of anchors in $\mathcal{A}$, and $N_c$ is the number of samples in $C$. While, IGNO's complexity, which requires multiplicative operations and matrix-solving computations, is larger than $(24\times N_a+30)\times N_{iter}$, where $N_{iter}$ is the iteration count. The PPPSC enables parallel operations on the heatmaps within a batch, and facilitates parallel processing on both the encoding and decoding steps on the GPU. The PPPSC's complexity is much less than IGNO's. The following experiment result in Table \ref{tab:fps} also prove this conclusion.


\section{Experiments}
\label{sec:experiments}

\subsection{Dataset and Metric}
\label{sec:dataset}
\subsubsection{Datasets} 
Five landmark detection datasets are employed for a comprehensive evaluation of the POPoS framework:
\noindent
\begin{itemize}[leftmargin=*]
\item \textbf{COFW}~\cite{COFW_ICCV2013} comprises 1,345 training and 507 testing images, each with 29 annotated landmarks. This dataset emphasizes scenes with faces under occlusion.
\item \textbf{AFLW}~\cite{Kostinger_AFLW_ICCVW11} includes a diverse set of approximately 25,000 face images from Flickr, annotated with up to 21 landmarks per image, showcasing wide variations in appearance and environmental conditions.
\item \textbf{COCO-WholeBody}~\cite{Jin_COCOWholebody_ECCV20} offers a collection of over 200,000 labeled images and 250,000 instances across 133 keypoint categories, including comprehensive annotations for face landmarks.
\item \textbf{300W}~\cite{Sagonas_300W_IVC16} consists of approximately 3,937 face images, each annotated with 68 facial landmarks, encompassing a broad spectrum of identities, expressions, and lighting conditions.
\item \textbf{WFLW}~\cite{Wu_LaB_CVPR18} is a widely-used dataset in facial landmark detection, containing 10,000 images (7,500 for training, 2,500 for testing), that each image is annotated with 98 landmarks.
\end{itemize}

\subsection{Evaluation Metric}~Following standard practice in facial landmark detection~\cite{bao2023keyposs}, the Normalized Mean Error (NME) is used as evaluation metric, which is defined as $\frac{1}{N_k}\sum_{i=1}^N\frac{\|\beta_{i} - \hat{\beta}_{i}\|_2}{d}$,
where $N_k$ is the number of landmarks, $\beta$ and $\hat{\beta}$ represent the ground truth and predicted landmarks, respectively, and $d$ is the normalization factor, typically the interpupillary distance.

\subsection{Implementation Details}
\label{sec:implementation_details}
Our proposed POPoS framework builds upon the KeyPosS and MMpose\footnote{https://github.com/open-mmlab/mmpose} frameworks. Under same condition with KeyPosS’s preprocessing and augmentation techniques, we evaluate the performance at varying heatmap resolutions: $64\times64$, $32\times32$, $16\times16$, $8\times8$, and $4\times4$ pixels. The backbone network is a lightweight variant of HRNet~\cite{Sun_HRNet_CVPR19}, which is pre-trained on ImageNet~\cite{ILSVRC15} and fine-tuned using the Adam optimizer~\cite{Kingma_Adam_ICLR15} with a linear-step decay learning rate schedule. The initial learning rate is set at 2.0e-3, and gradually reducing to 1.0e-5 over 100 epochs. The training is performed on a 4 × NVIDIA 3090 GPU server, with a batch size 32. In the decoding stage of POPoS, the key hyperparameters, such as the sampling range (w=1) and the number of anchors (K=10), are determined through training on the 300W~\cite{Sagonas_300W_IVC16} dataset.

\begin{figure*}[!t] 
    \centering
    \includegraphics[width=.9\linewidth]{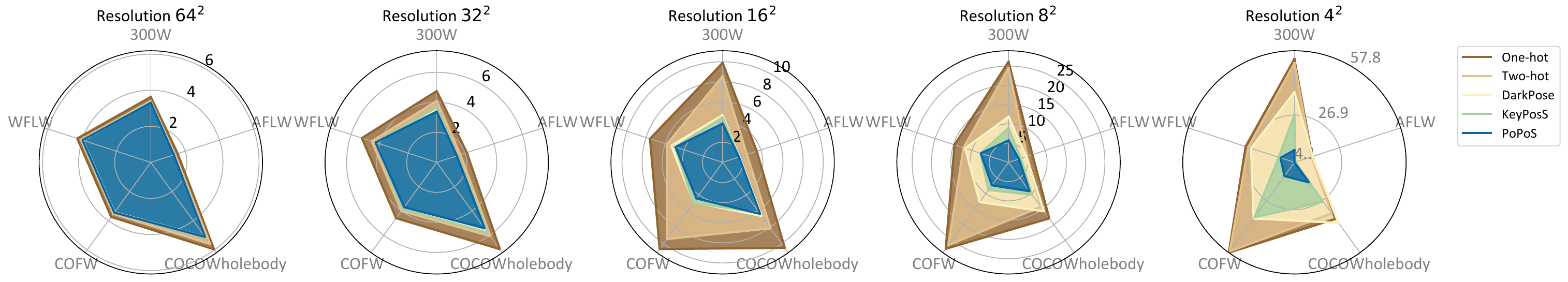}
    \vspace{-0.5em}
    \caption{Keypoint detection accuracy at heatmap resolutions of \( 64 \times 64 \), \( 32 \times 32 \), \( 16 \times 16 \), \( 8 \times 8 \), and \( 4 \times 4 \).~[The smaller radar map denotes the better performance]}
    \label{fig:low-resultion}
     \vspace{-1.5em}
\end{figure*}

\begin{table}[!t] 
\scriptsize
\centering
        \caption{Comparison with the State-of-the-Art methods. The results are in NME (\%). The best results are highlighted with bold text font. * means the result of IGNO solution. }
        \vspace{-0.1in}
    \setlength{\tabcolsep}{0.1mm}
    \begin{tabular}[t]{c|c|c|c|c|c|c}
        \toprule
        Method                                     & WFLW & 300W & AFLW    & COFW & Para. & GFlops\\
        \midrule
        TSR \cite{Lv_TSR_CVPR17}                   & -                                        & 4.99                                          & 2.17                                               & -                                        & -          & -\\ 
        Wing  \shortcite{Feng_Wingloss_CVPR18}          & -                                        &  3.60                                         & 1.47                                               & -                                        & 91.0M      & 5.5\\ 
        ODN \shortcite{Zhu_OAD_CVPR19}                  & -                                        & 4.17                                          & 1.63                                               & 5.30                                     & -          & -\\ 
        DeCaFa \shortcite{Dapogny_DeCaFA_ICCV19}        & 4.62                                     & 3.39                                          & -                                                  & -                                        & 10M        & -\\ 
        DAG \shortcite{Li_DAG_ECCV20}                   & 4.21                                     & 3.04                                          & -                                                  & 4.22                                     & -          & -\\
        \midrule
        AWing \shortcite{Wang_AWingLoss_ICCV19}         & 4.36                                     & 3.07                                          & 1.53                                               & 4.94                                     & 24.1M      & 26.7\\ 
        AVS \shortcite{Qian_AVS_ICCV19}                 & 4.39                                     & 3.86                                          & 1.86                                               & 4.43                                     & 28.3M      & 2.4\\ 
        ADA \shortcite{Chandran_attention_CVPR2020}     & -                                        & 3.50                                          & -                                                  & -                                        & -          & -\\ 
        LUVLi \shortcite{Kumar_LUVLi_CVPR20}            & 4.37                                     & 3.23                                          & 1.39                                               & -                                        & -          & -\\ 
        PIPNet-18 \cite{Jin_PIPNet_IJCV21}         & 4.57                                     & 3.36                                          & 1.48                                               & -                                        & 12.0M      & 2.4\\ 
        PIPNet-101 \cite{Jin_PIPNet_IJCV21}        & 4.31                                     & 3.19                                          & 1.42                                               & -                                        & 45.7M      & 10.5\\ 
        DTLD \cite{Li_transformer_CVPR22}          & 4.08                                     & {2.96}                                        & 1.38                                               &  -                                       & -          & -\\ 
        RePFormer \cite{li2022repformer}           & 4.11                                     & 3.01                                          & 1.43                                               & -                                        & -          & -\\ 
        SLPT \cite{xia2022sparse}                  & 4.14                                     & 3.17                                          & -                                                  & -                                        & 9.98M      & -\\ 
        EF-3-ACR \cite{fard2022acr}                & -                                        & 3.75                                          & -                                                  & 3.47                                     & -          & -\\ 
        ADNet-FE5 \cite{huang2023freeenricher}     & 4.1                                      & 2.87                                          & -                                                  & -                                        & -          & -\\ 
        ResNet50-FE5 \cite{huang2023freeenricher}  & -                                        & 4.39                                          & -                                                  & -                                        & -          & -\\ 
        HRNet-FE5 \cite{huang2023freeenricher}     & -                                        & 3.46                                          & -                                                  & -                                        & -          & -\\ 
        STAR Loss \cite{zhou2023star}              & 4.02                                     & \textbf{ 2.87}                                & -                                                  & 4.62                                     & 13.37M     & -\\ 
        RHT-R \cite{wan2023precise}                & 4.01                                     & 3.46                                          & 1.99                                               & 4.42                                     & -          & -\\  

        \midrule
        POPos (ours, $32^2$)                       & {4.28}                            & 3.38                                          & 1.43                                               & {3.8}                             & 9.7M       & \textbf{1.2}\\
        POPos (ours, $64^2$)                       & \textbf{3.95*}                           & 3.28*                                         & \textbf{1.34}                                      & \textbf{3.44}                            & 9.7M       & 4.7\\
        \bottomrule
    \end{tabular}
    \label{tab:SOTA}
\vspace{-1em}
\end{table}
\begin{table}[!t]   
\scriptsize
    \centering
 \caption{FPS of existing decoding methods and PoPoS}
 \vspace{-0.1in}
    \begin{tabular}{c|c|c}
        \hline
        & FPS  & Cost \\
        \hline
        One-hot & 2819 & 0.36ms \\
        \hline
        Two-hot~\cite{Newell_Hourglass_ECCV16} & 1394 & 0.72ms \\
        \hline
        DarkPose~\cite{Zhang_darkpose_CVPR20} & 186.2 & 5.3ms \\
        \hline
        KeyPosS~\cite{bao2023keyposs} & 96.5 & 10.34ms \\
        \hline
        IGNO for PRM  & 2.5 & 405.06ms \\
        \hline
        POPoS & 1301 & {0.76ms} \\
        \hline
    \end{tabular}
\vspace{-0.2in}
    \label{tab:fps}
\end{table}
\subsection{Comparison with State-of-the-Art Methods}
\label{sec:ext-sota}

 The effectiveness of the Parallel Optimal Position Search (POPoS) is evaluated against state-of-the-art methods across five pivotal datasets, detailed in Table~\ref{tab:SOTA}. The NME of POPoS is less than the traditional coordinate regression-based method and other heatmap-based methods.~We think that coordinate regression-based methods (TSR,Wing, et al) compromise spatial detail retention, thereby affecting precision in high-resolution scenarios. POPoS's exceptional performance, particularly on the WFLW, AFLW, COFW, and COCO-Wholebody datasets, emphasizes its improved capability in addressing encoding and decoding errors.

 Notably, POPoS achives impressive performance in WFLW, AFLW, and COFW with fewer parameters and higher computational efficiency. On the 300W dataset, POPoS reaches a competitive result Normalized Mean Error (NME) of 3.28\%, highlighting the complexity of the dataset and the challenges with limited training data. Notice that, the NME value of the COCO-Wholebody dataset is only used to analyze the decoding precious of various heatmap resolutions, as seen in Fig.~\ref{fig:low-resultion}. 

\subsection{Comparison with Decoding Schemes} The key problem addressed in this article is the verification of decoding accuracy at different resolutions. The decoding effects of five different methods were compared across five datasets under various heatmap resolution settings. As shown in Fig.~\ref{fig:low-resultion}, a distinctive advantage of POPoS is found in the integration of the Iterative Gauss–Newton Optimization (IGNO) approximation method with a strategic sampling approach. This integration is especially beneficial in low-resolution heatmaps settings, where high-quality results are delivered efficiently by POPoS.

Additionally, to evaluate the efficiency of our proposed POPoS on heatmap decoding, the processing speed of different decoding methods, with the metric of Frame per second~(FPS), is calculated. Although the current heatmap inference speed reaches 200 FPS~\cite{Jin_PIPNet_IJCV21}, slow heatmap decoding could potentially become a bottleneck. The decoding speed of five methods, one-hot, two-hot, Dark~\cite{Zhang_darkpose_CVPR20}, KeyPosS~\cite{bao2023keyposs}, and POPoS are compared in Table \ref{tab:fps}. The FPS for \(256 \times 256\)  input images with \(64 \times 64\) heatmaps was measured to assess decoding speed. It is clear that the POPoS is 15 times faster than KeyPos, 6 times faster than Dark~\cite{Zhang_darkpose_CVPR20}, and comparable in speed to Two-Hot. Although slower than One-Hot, POPoS has significantly improved accuracy and meets real-time processing requirements.

\begin{figure}[!t]
    \centering
    \vspace{-0.05in}
    \includegraphics[width=0.9\linewidth]{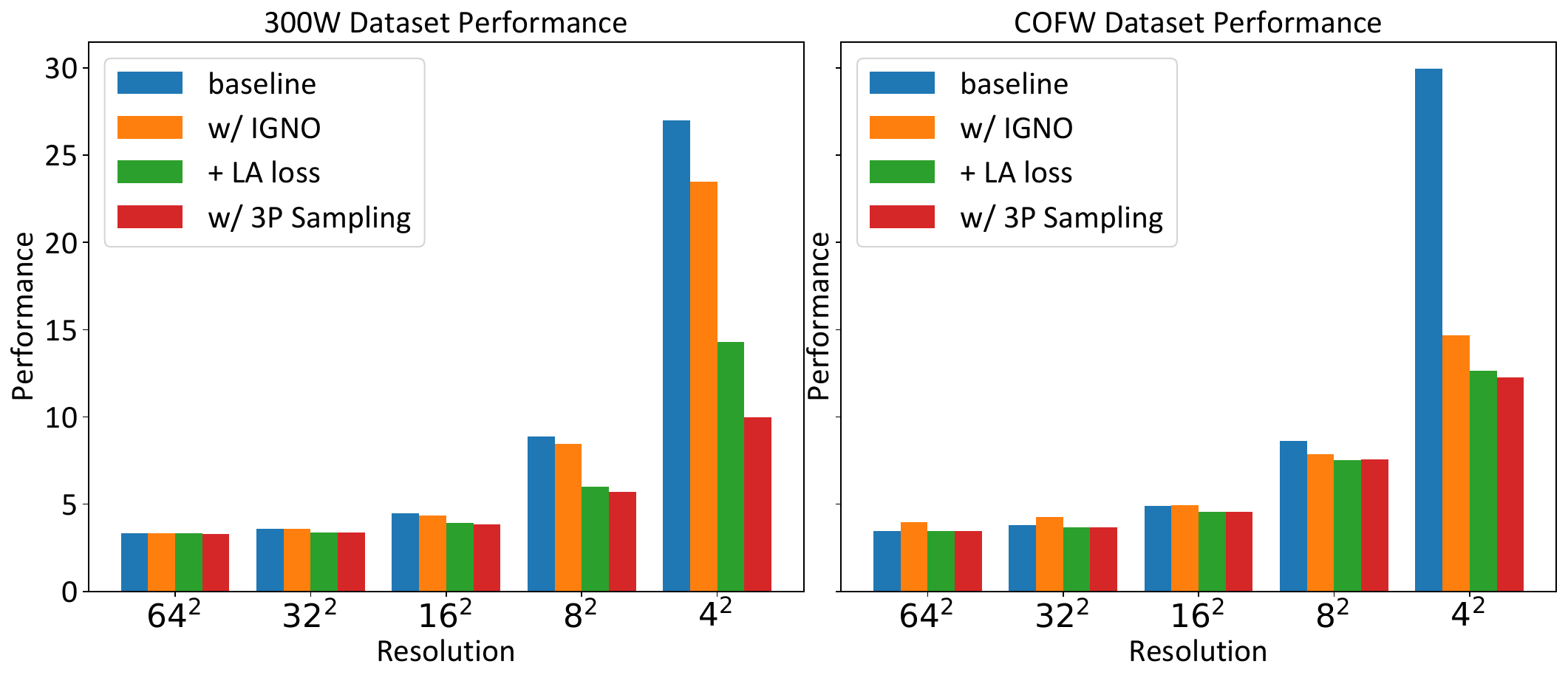}
    \vspace{-0.1in}
    \caption{Effect of Gauss-Newton Optimization~(IGNO), Multilateration Anchor~(MA) loss, and Potential Position Parallel Sampling
and Computing~(PPPSC).}
    \label{fig:ablation}
    \vspace{-0.2in}
\end{figure}

\begin{figure*}[!th]
    \centering
    \begin{minipage}{0.24\linewidth}
        \centering
        \subfloat[]{\label{fig:vis-sample_area1}
        \includegraphics[width=0.9\linewidth]{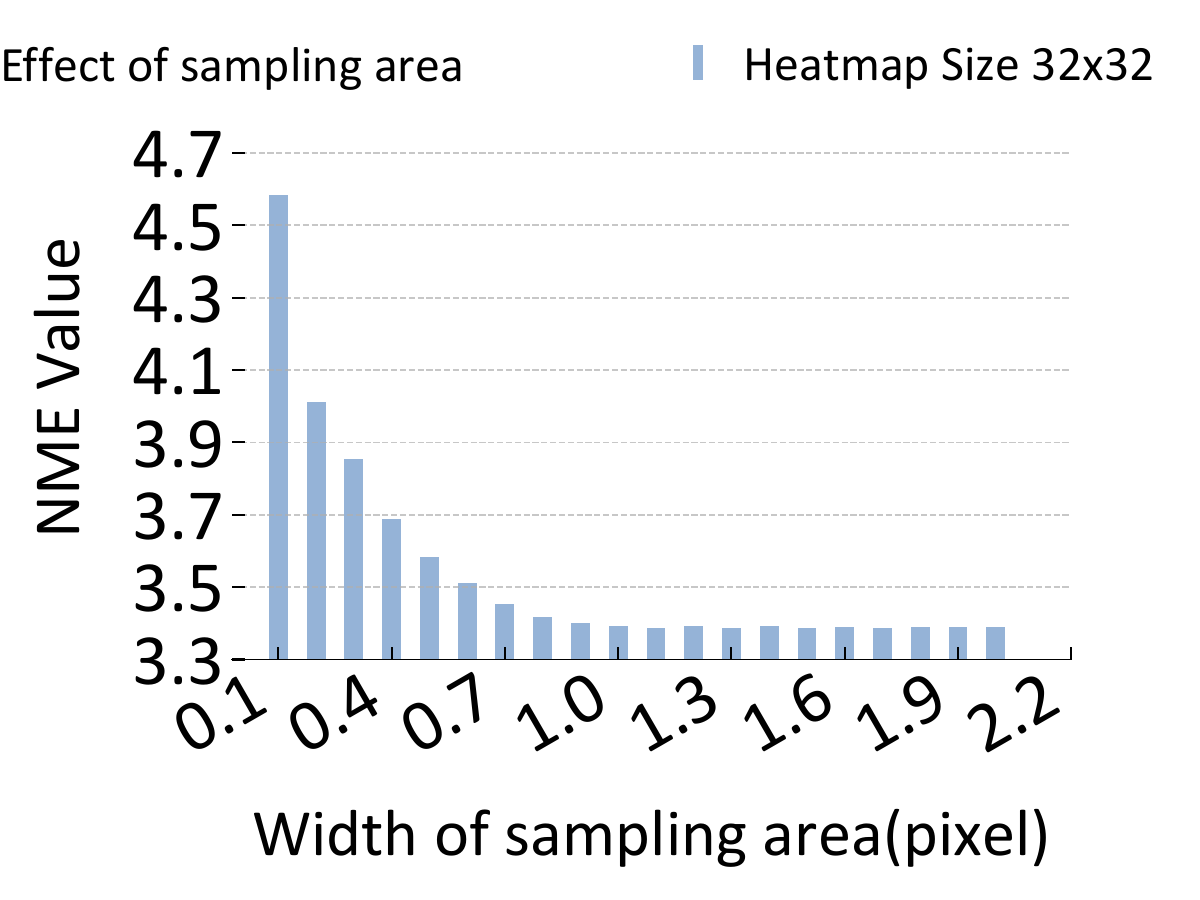}}
    \end{minipage}
    \hfill
    \begin{minipage}{0.24\linewidth}
        \centering
        \subfloat[]{\label{fig:vis-sample_num}
        \includegraphics[width=0.95\linewidth]{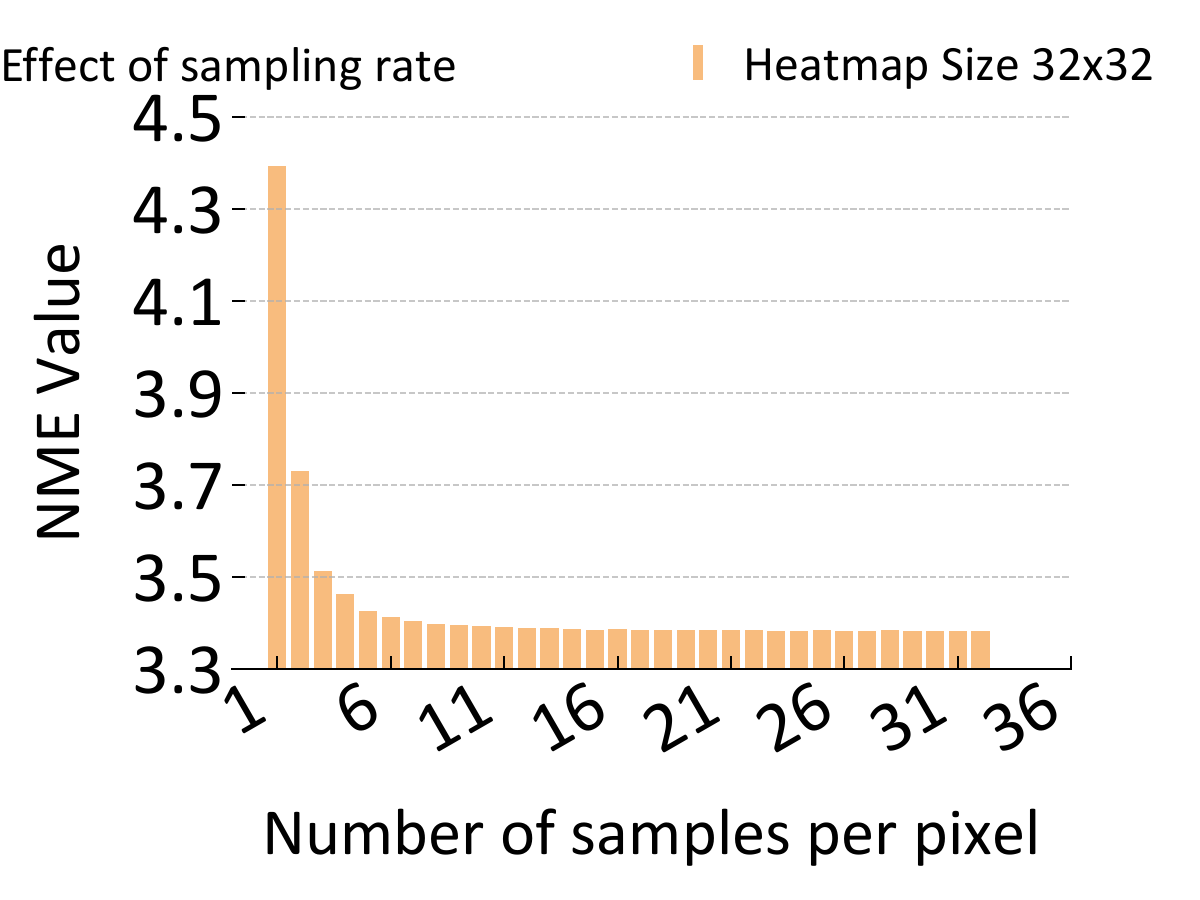}}
    \end{minipage}
    \hfill
    \begin{minipage}{0.24\linewidth}
        \centering
        \subfloat[]{\label{fig:vis-anchor_num}
        \includegraphics[width=0.95\linewidth]{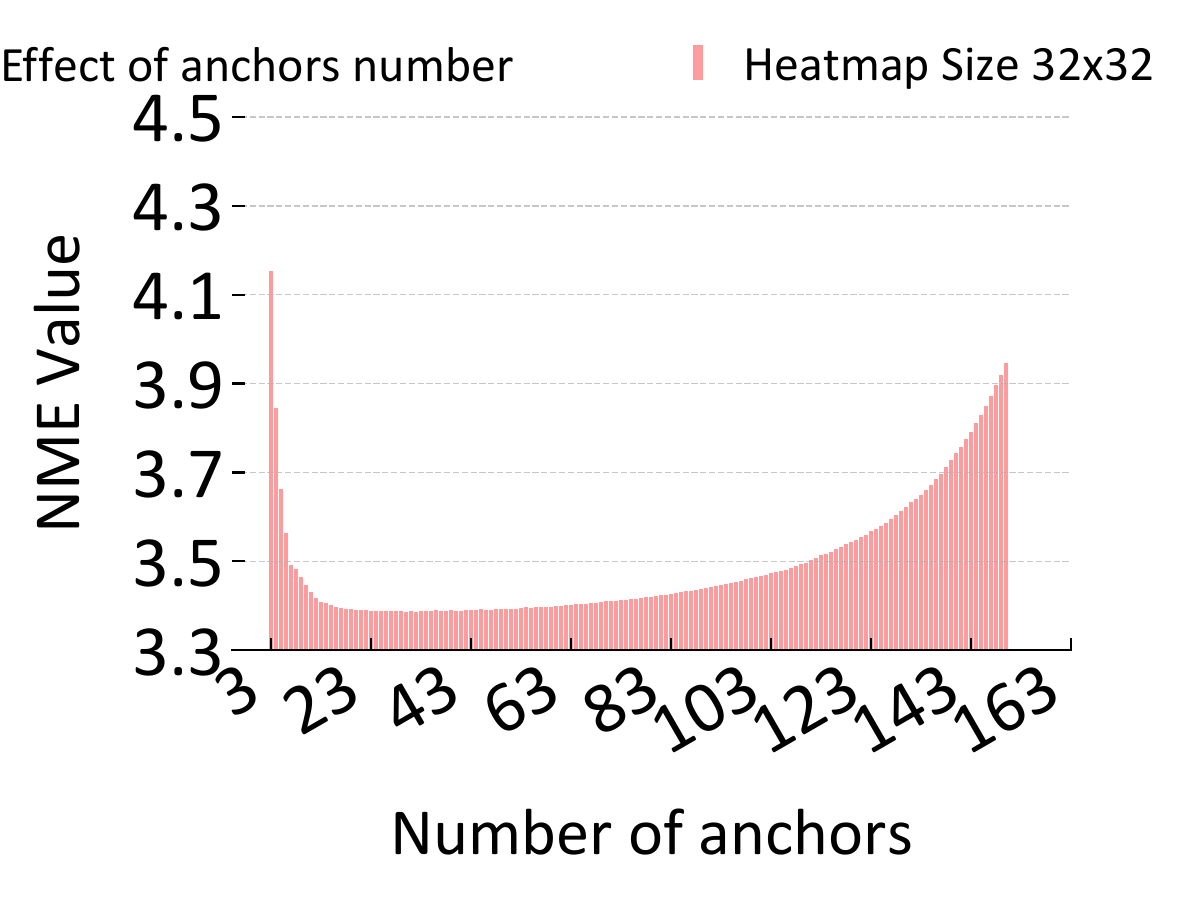}}
    \end{minipage}
    \hfill
    \begin{minipage}{0.24\linewidth}
        \centering
        \subfloat[]{\label{fig:vis-loss_weight}
        \includegraphics[width=0.95\linewidth]{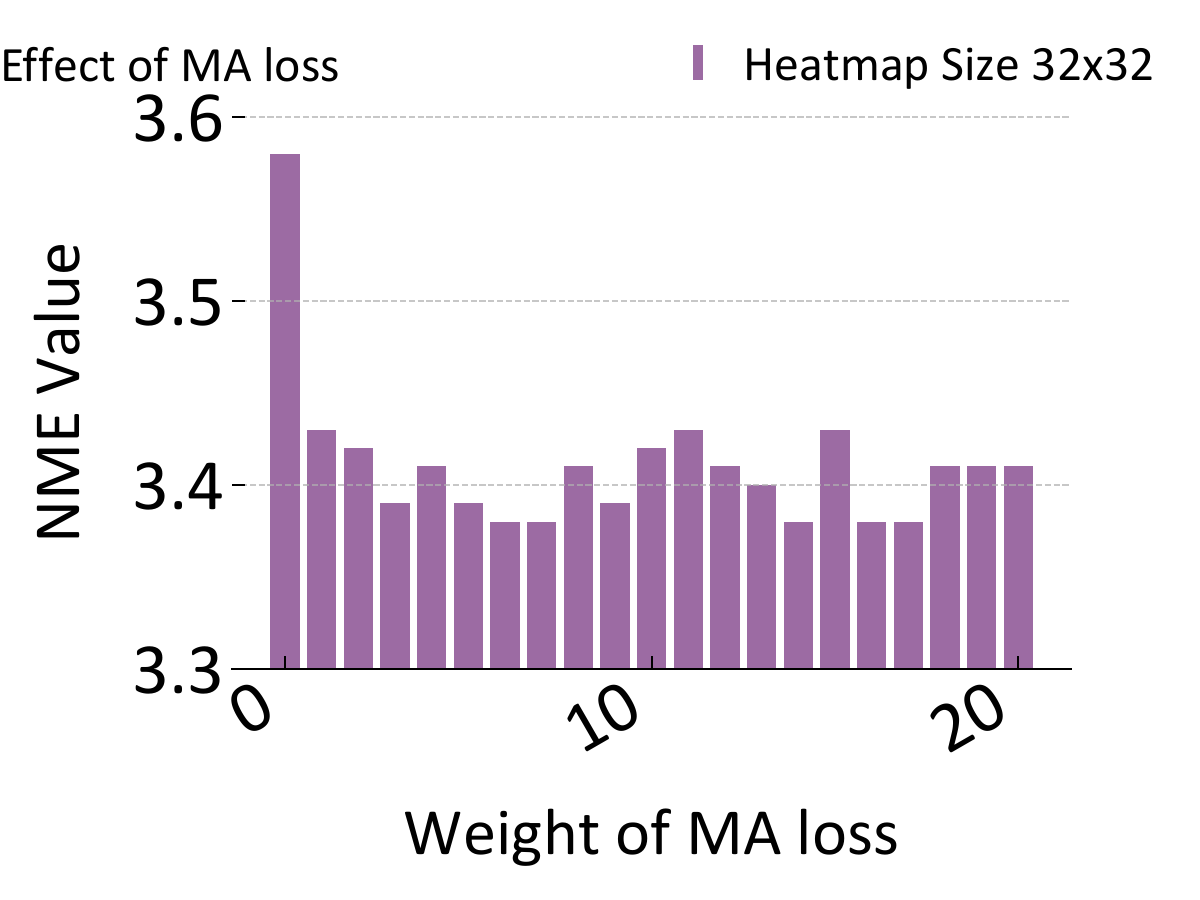}}
    \end{minipage}
    \vspace{-0.1in}
    \caption{Detailed analysis of POPoS performance factors: (a) Effect of sampling area (b) Effect of sampling rate $\tau$ (c) Anchor number and (d) MA loss weight, each illustrating their impact on the accuracy of predictions.}
    \label{fig:vis-anchor-and-loss}
\vspace{-0.1in}
\end{figure*}
\begin{figure*}[!th]
\centering
\begin{center}
    \includegraphics[width=.8\linewidth]{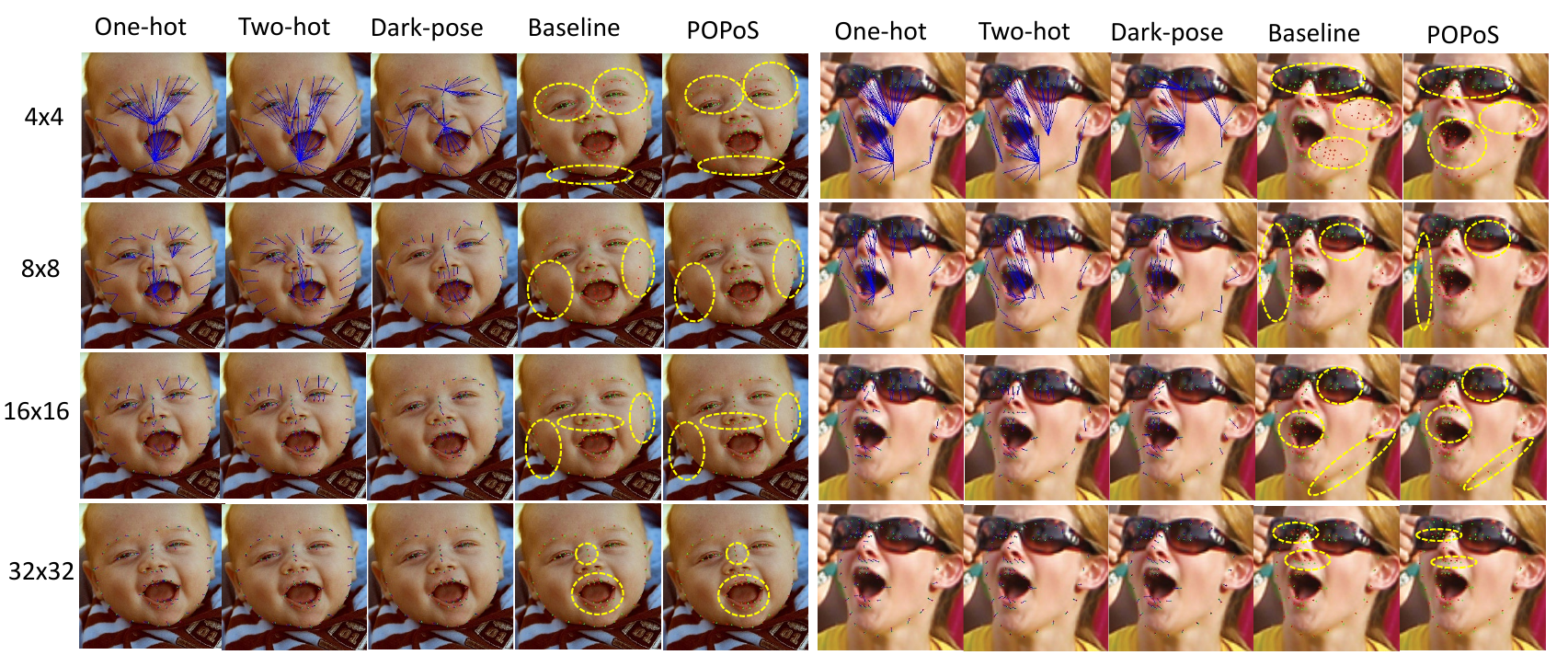}
\end{center}
\vspace{-0.1in}
\caption{Visualization of POPoS's performance in low-resolution heatmaps settings: Comparative analysis with one-hot, two-hot, dark-pose, and KeyPosS methods. Ground truth is represented by \textcolor{green}{green dots}, predictions by \textcolor{red}{red dots}, discrepancies by \textcolor{blue}{blue lines}, and significant inaccuracies are circled in \textcolor{yellow}{yellow}.~[Best viewed in color and at a higher zoom level]}
\label{fig:vis-landmarks}
\vspace{-0.2in}
\end{figure*}

\subsection{Ablation Studies}
\label{sec:ablation}
This section offers an in-depth analysis of the effects of the proposed modules and hyperparameter settings.

\subsubsection{Effect of Pseudo-Range Multilateration}
\label{sec:effect_prm}
The effectiveness of Iterative Gauss–Newton Optimization (IGNO) for pseudo-range multilateration (PRM) was evaluated by integrating it into the baseline model, replacing the least squares approximation. The number of iterations was capped at 20 or halted upon convergence. KeyPosS~\cite{bao2023keyposs} served as the baseline. Improvements were observed in the 300W and COFW datasets, as illustrated in Fig.~\ref{fig:ablation}. These enhancements are attributed to a reduction in prediction errors achieved through an increased number of anchors.

\subsubsection{Effect of MA Loss}
\label{sec:effect_ma_loss}
The impact of Multilateration Anchor (MA) loss on mitigating encoding errors was assessed by retraining the HRNet network across various resolutions. The integration of the multilateration anchor loss resulted in significant improvements in POPoS across all datasets, as depicted in Fig.~\ref{fig:ablation}. Specifically, in the \(4 \times 4\) resolution heatmap, prediction errors decreased by 47\% for the 300W dataset and by 58\% for the COFW dataset.

\subsubsection{Effect of PPPSC}
\label{sec:effect_pppsc}
Potential Position Parallel Sampling and Computing (PPPSC) was evaluated on five datasets at varying resolutions, as presented in Fig.~\ref{fig:ablation}. This approach significantly enhanced the stability and performance of the IGN Optimization algorithm. Ablation experiments on the sampling range $(h,w)$ and sampling frequency $\tau$ were conducted using the 300W dataset at a resolution of \(32 \times 32\). The results, shown in Fig.~\ref{fig:vis-anchor-and-loss}, indicated that the optimal sample area values lie within 1-2 pixels around the maximum value. Increasing the sampling range did not improve decoding accuracy, suggesting that some errors originate in the encoding stage and cannot be corrected by decoding adjustments. Additionally, ablation studies on the sample rate $\tau$ demonstrated that optimal performance is achieved with approximately 10 sampling points per pixel length, equivalent to about 1.2 points per original image pixel, as shown in Fig.~\ref{fig:vis-sample_area1}. Further increases in the number of sampling points resulted in negligible improvements.

\subsubsection{Ablation Study on Hyperparameter Settings}
\label{sec:ablation_hyperparams}
The ablation study on the number of anchors \(N_a\) in the decoding step, as shown in Fig.~\ref{fig:vis-anchor_num}, indicates that too few anchors result in inaccurate predictions, while an excessive number of anchors introduces larger calculation errors. For a \(32 \times 32\) resolution heatmap, the optimal number of anchors was determined to be 25. Additionally, an ablation experiment on the weight of the local anchor loss in the 300W dataset (Fig.~\ref{fig:vis-loss_weight}) demonstrated consistent improvements in prediction accuracy across various weight ratios, with the most effective ratio being 6. This finding highlights the significant impact of encoding accuracy errors on pseudo-range multilateration and the efficacy of the local anchor loss in mitigating convergence issues.

\subsection{Visualization Analysis}
\label{sec:ext:visul}
To evaluates POPoS's adaptability in practical applications, we conducted a comprehensive visualization analysis comparing it with traditional methods such as one-hot, two-hot, DarkPose, and KeyPosS. Focusing on challenging low-resolution heatmaps scenarios depicted in Fig.~\ref{fig:vis-landmarks}, the analysis revealed that conventional methods struggle with precision at heatmap resolutions of $4\times4$ and $8\times8$. In contrast, POPoS demonstrates superior performance starting from $4\times4$ and maintains high accuracy at higher resolutions. Additionally, POPoS effectively mitigates localization errors across all tested resolutions, as clearly shown in Fig.~\ref{fig:vis-landmarks}. These visualizations confirm POPoS's robustness, highlighting its ability to optimize encoding and decoding processes effectively across diverse application scenarios.

\section{Conclusion}
\label{sec:conclusion}
We introduce Parallel Optimal Position Search (POPoS), a heatmap-based framework that effectively balances accuracy and efficiency in facial landmark detection (FLD). POPoS employs the \textit{Multilateration Anchor Loss} during encoding to reduce heatmap prediction errors by enhancing anchor point accuracy. In decoding, an optimization-based multilateration method surpasses traditional approaches in landmark localization. Additionally, the \textit{Equivalent Single-Step Parallel Computation} algorithm significantly improves computational efficiency, enabling real-time FLD applications. Extensive evaluations on five benchmark datasets demonstrate that POPoS consistently outperforms existing FLD methods, particularly in low-resolution scenarios. 

\section*{Acknowledgments}
Zhi-Qi Cheng’s research for this project was supported in part by the US Department of Transportation, Office of the Assistant Secretary for Research and Technology, under the University Transportation Center Program (Federal Grant Number 69A3551747111). This work was also supported in part by the National Natural Science Foundation of China (Grant No. 62372387) and the Key R\&D Program of Guangxi Zhuang Autonomous Region, China (Grant Nos. AB22080038 and AB22080039).

\bibliography{aaai25}

\end{document}